\documentclass[10pt,twocolumn,letterpaper]{article}

\usepackage[pagenumbers]{cvpr} 

\definecolor{cvprblue}{rgb}{0.21,0.49,0.74}
\usepackage[pagebackref,breaklinks,colorlinks,allcolors=cvprblue]{hyperref}

\usepackage{bm}
\usepackage{multirow}
\usepackage{graphicx,verbatim}
\usepackage{amssymb}
\usepackage{amsmath}

\def\paperID{*****} 
\def\confName{CVPR}
\def\confYear{2026}

\title{ASBA: A-line State Space Model and B-line Attention for Sparse Optical Doppler Tomography Reconstruction}

\author{Zhenghong Li \quad 
Wensheng Cheng \quad 
Congwu Du \quad
Yingtian Pan \quad 
Zhaozheng Yin \quad 
Haibin Ling$^{*}$ \\
[2mm]
Stony Brook University
}

\begin{document}
\maketitle

\let\thefootnote\relax\footnotetext{$^*$ Work was done while with Stony Brook University.}

\begin{abstract}
Optical Doppler Tomography (ODT) is an emerging blood flow analysis technique. A 2D ODT image (B-scan) is generated by sequentially acquiring 1D depth-resolved raw A-scans (A-line) along the lateral axis (B-line), followed by Doppler phase-subtraction analysis. To ensure high-fidelity B-scan images, current practices rely on dense sampling, which prolongs scanning time, increases storage demands, and limits the capture of rapid blood flow dynamics. Recent studies have explored sparse sampling of raw A-scans to alleviate these limitations, but their effectiveness is hindered by the conservative sampling rates and the uniform modeling of flow and background signals. In this study, we introduce a novel blood flow-aware network, named \textbf{ASBA} (\textbf{A}-line ROI \textbf{S}tate space model and \textbf{B}-line phase \textbf{A}ttention), to reconstruct ODT images from highly sparsely sampled raw A-scans. Specifically, we propose an A-line ROI state space model to extract sparsely distributed flow features along the A-line, and a B-line phase attention to capture long-range flow signals along each B-line based on phase difference. Moreover, we introduce a flow-aware weighted loss function that encourages the network to prioritize the accurate reconstruction of flow signals. Extensive experiments on real animal data demonstrate that the proposed approach clearly outperforms existing state-of-the-art reconstruction methods. 
\end{abstract}

\section{Introduction}
\label{sec:intro}
Optical Doppler Tomography (ODT) is an emerging technique for analyzing blood flow~\cite{leitgeb2014doppler}. ODT measures blood flow speed by exploiting the Doppler phase shift induced by moving red blood cells within vessels~\cite{chen2008doppler}. It offers high-resolution high-contrast tomographic flow imaging, making it suitable for many clinical applications such as vascular disease monitoring~\cite{li2020deep}. Despite its promise in blood flow speed estimation, current ODT technique requires dense spatial sampling to accurately capture flow dynamics, particularly to detect slow-moving red blood cells in small vessels~\cite{wang2007three}. This dense scanning strategy results in long acquisition times and substantial data storage demands. Moreover, the long scanning time further limits the detection of high dynamic changes in blood flow caused by stimulation.

\begin{figure}[!t]
\centering
\includegraphics[width=.9\linewidth]{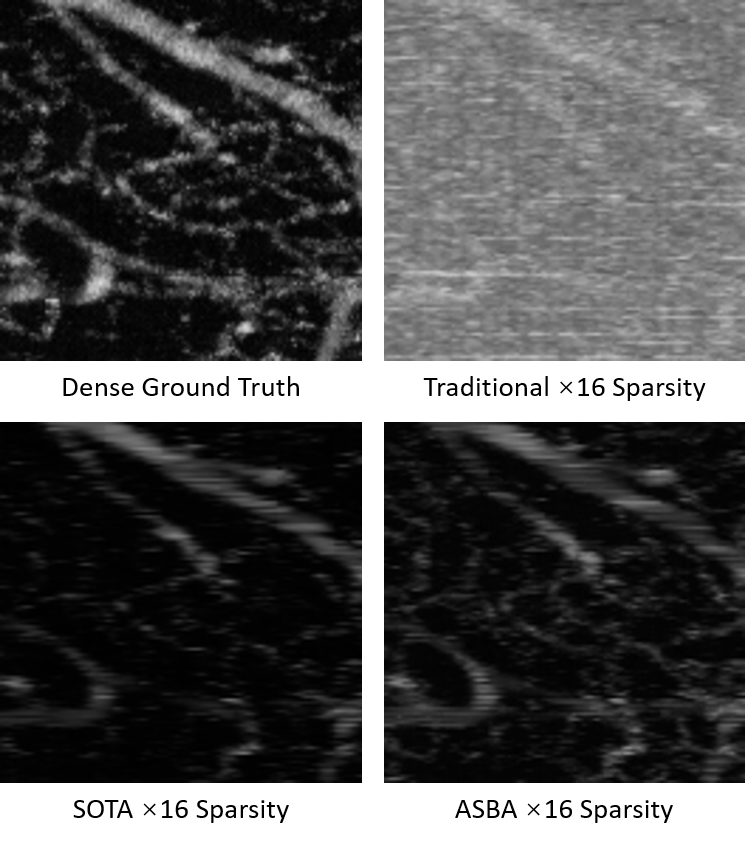}
    \caption{Examples of sparse ODT reconstruction. Due to big gaps among sampled A-lines at high sparsity, traditional pipeline~\cite{zhao2000phase} that estimates flow speed directly from phase differences of adjacent A-lines fails to reconstruct flow images. The state-of-the-art (SOTA) sparse ODT reconstruction method~\cite{li2025sparse} also loses most weak flow signals in small vessels, while our proposed ASBA effectively preserves most flows in the reconstructed image.}
\label{fig:receg}
\end{figure}

In typical ODT reconstruction, the fundamental scan, named raw A-scan, encodes the depth information (A-line) in the spectral domain. A 2D raw B-scan is acquired by laterally sliding the sensor along the width (B-line). To retrieve depth information, 1D IFFT is applied to each A-line, decoding the depth-resolved magnitude and phase responses in the spatial domain. Phase differences between adjacent A-lines are used to estimate the blood flow speed map~\cite{leitgeb2014doppler}, and magnitude to suppress background noise. 

A recent study~\cite{li2025sparse} shows the potential of reconstructing ODT images from \textbf{sparsely} sampled raw A-scans. However, this approach is limited in both scanning strategies and modeling methodologies. First, it employs conservative sparse sampling rates ($\times 2$ or $\times 4$ sparsity), failing to substantially shorten scanning times. Second, it uniformly models the magnitude and phase signals, as well as the flow and background regions, without explicit distinction. Given that magnitude and phase play distinct roles in ODT reconstruction, and considering that background regions comprise, on average, more than 94\% of the B-scan images in our experiments, such undifferentiated modeling strategies are suboptimal for accurate flow reconstruction. 

It is also worth noting that, despite the superficial resemblance between sparse ODT reconstruction and single-image super-resolution (SISR)~\cite{liang2021swinir,chen2023activating,li2023efficient,zhou2023srformer,chen2023dual,ray2024cfat,zhang2024transcending,guo2024mambair,guo2025mambairv2,li2025mair}, existing SISR methods are inappropriate for sparse ODT reconstruction. Besides the above mentioned limitations, this is also primarily due to the unique characteristics of ODT phase data—where phase differences, rather than individual values, are critical—and the fundamentally different sampling strategy, which involves directly skipping A-lines rather than aggregating neighboring pixels.

To more effectively reduce scanning time and memory requirements, we propose reconstructing ODT B-scan images from \textbf{highly sparsely} sampled raw A-scans. Given that flow speed estimation in traditional ODT relies on the phase difference between adjacent raw A-scans~\cite{zhao2000phase}, a high sparsity implies insufficient information for conventional flow reconstruction for each A-scan. As shown in Figure~\ref{fig:receg}, such an aggressive sampling strategy introduces substantial noise and artifacts, making vessels and flows invisible in traditionally reconstructed ODT. Therefore, given the inherent sparsity of flow regions in B-scans and the distinctive features of phase signals, a dedicated sparse reconstruction model that explicitly emphasizes flow modeling is essential.

\begin{figure*}[!t]
\centering
\includegraphics[width=.95\linewidth]{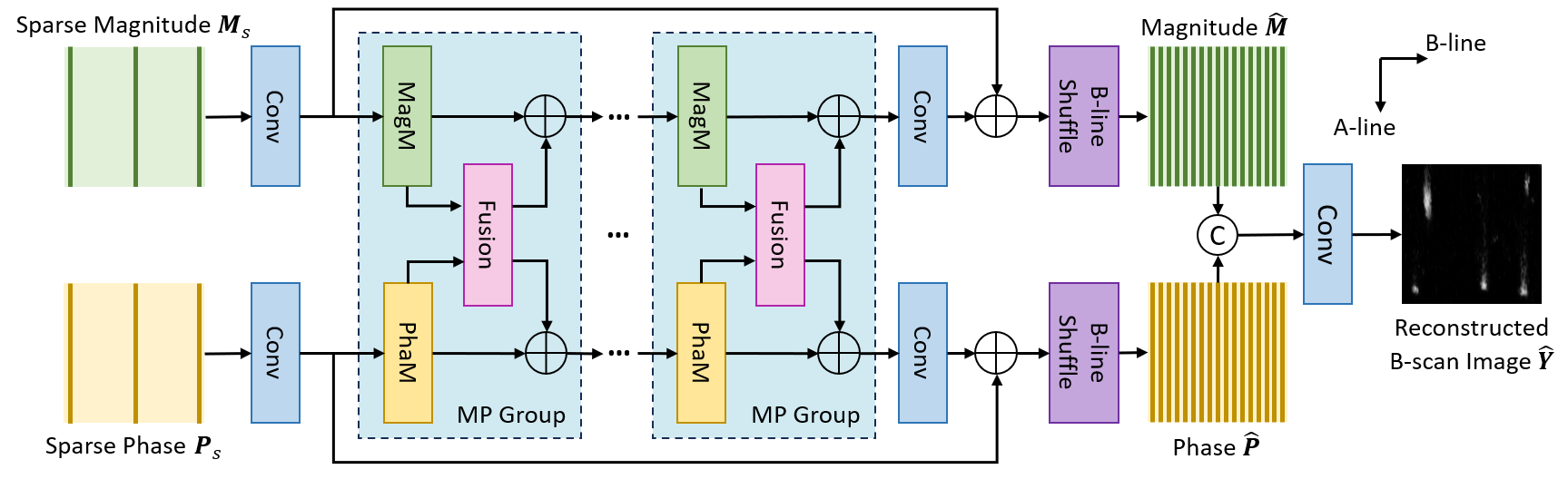}
    \caption{Our sparse ODT reconstruction system. The raw B-scan is highly downsampled with a large stride along the B-line. The sparse magnitude and phase responses are forwarded to the proposed ASBA network to reconstruct the B-scan image.}
\label{fig:pipeline}
\end{figure*}

In this work, we propose a flow-aware reconstruction network named \textbf{ASBA} (\textbf{A}-line ROI \textbf{S}tate Space Model and \textbf{B}-line phase \textbf{A}ttention), aiming to address the modeling issues in previous works by focusing on flow information reconstruction from highly sparsely sampled ODT raw A-scans. The overall pipeline is demonstrated in Figure~\ref{fig:pipeline}. \textbf{First}, we design a two-branch framework to separately process the magnitude and phase information for their different attributions. Since phase information is the core of ODT reconstruction, we specifically focus on phase reconstruction in the network block design. \textbf{Second}, we propose an A-line Region-of-Interest (ROI) state space block to adaptively model the flow-relevant phase information along each A-line. As flow signals are sparsely distributed in depth and background signals contribute minimally to flow reconstruction, we dynamically modulate the input matrix of the State Space Model (SSM) to highlight the flow region and suppress the background for more effective state transition. \textbf{Third}, we introduce a B-line phase attention mechanism to model flow information along the B-line. Since it is the phase difference, rather than the phase itself, that indicates flow characteristics, explicitly incorporating phase differences into the attention computation better aligns the network with the intrinsic features of blood flow. \textbf{Finally}, we develop a flow-aware weighted loss function that adaptively assigns higher weights to the flow regions, encouraging the network to prioritize the reconstruction of flow features over the dominant but less informative background signals.

Based on the above designs, ASBA effectively captures flow information, enabling accurate reconstruction from highly sparsely sampled ODT raw B-scans, and hence significantly reducing data acquisition time and storage memory. Notably, the decreased scanning time facilitates the study of dynamic blood flow responses to various forms of stimulation. Experiments on both awake and anesthetized animal datasets demonstrate that our method has exhibited promising results in comparison with state-of-the-arts.

\section{Related Works}
\label{sec:related}
\textbf{Optical Doppler Tomography} (ODT) is an emerging non-invasive imaging approach for blood flow visualization~\cite{leitgeb2014doppler}. It employs phase difference of successive A-scans caused by Doppler frequency shift of moving red blood cells to estimate flow speed~\cite{chen1997noninvasive,izatt1997vivo,zhao2000phase}. However, because the sensitivity of ODT is highly dependent on the sampling interval~\cite{spaide2018optical}, dense sampling is typically required to detect slow blood flow~\cite{wang2007three}. Therefore, traditional ODT imaging requires a long scanning time and large storage memory. 

\textbf{Sparse ODT Reconstruction}.
Addressing the aforementioned challenge, a recent study in~\cite{li2025sparse} employs state space models for sparse ODT reconstruction. However, the approach is limited to conservative sparsity levels, thus failing to significantly address the scanning time and storage issue. Besides, it treats flow and background signals uniformly, overlooking the sparse nature of flow in ODT B-scans and the importance of phase difference for flow reconstruction.
Inspired by~\cite{li2025sparse} and meanwhile addressing its limitations, we propose a new solution for much more sparse ODT reconstruction.

Note that sparse reconstruction has been previously applied in other domains such as OCT, radar and MRI using compressive sensing~\cite{zhang2015survey,marques2018review,liu2010compressive,liu2011sparse,yang2016sparse,potter2008sparse,wei2010sparse}. However, due to domain discrepancies and differences in sampling methodologies, directly adapting these techniques to sparse ODT reconstruction remains non-trivial or inappropriate.

\textbf{State Space Models} (SSMs)~\cite{kalman1960new} were originally proposed in the field of control, describing the transformation of state spaces. Recently, a new SSM named Mamba~\cite{gu2023mamba} was introduced, extending early deep learning-based SSMs~\cite{gu2021efficiently,gu2021combining} by using input data to predict state matrices to facilitate selective sequence modeling, improving long-range dependencies with linear complexity. Due to these advantages, Mamba has been widely adopted for various vision tasks~\cite{zhuvision,liu2024vmamba,huang2024localmamba,chen2024rsmamba,zhao2024rs,zou2024freqmamba,guo2024mambair,ma2024u,li2025mair,guo2025mambairv2}.

Although Mamba can selectively process sequences, its intrinsic selective ability is not effective for sparse ODT reconstruction, given that the majority of an ODT B-scan is the background that contributes little to flow modeling.

\begin{figure*}[!t]
\centering
\includegraphics[width=.95\linewidth]{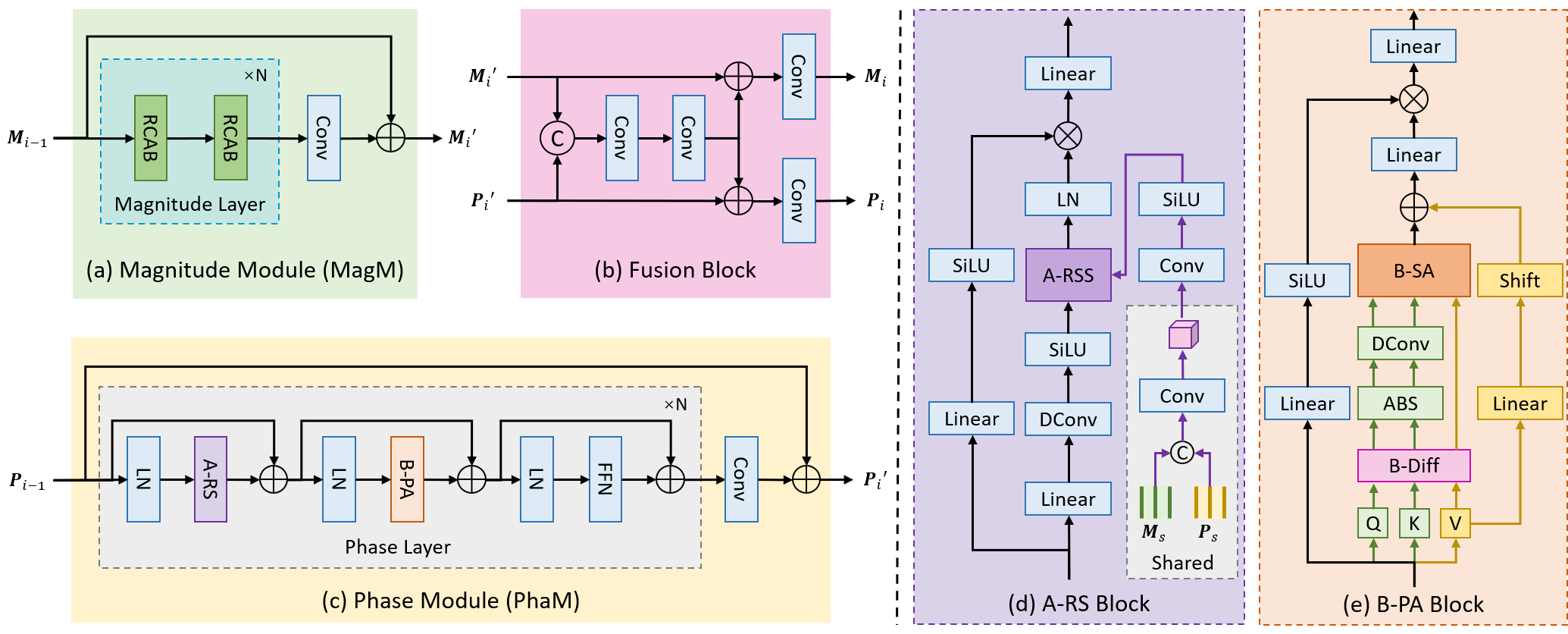}
    \caption{Network Blocks. (a) Magnitude Module, (b) Fusion Block, (c) Phase Module, (d) A-line ROI State Space Model (A-RS) Block, and (e) B-line Phase Attention (B-PA) Block. }
\label{fig:block}
\end{figure*}

\section{Method}
\label{sec:method}
\subsection{Problem Formulation}
\label{sec:formulation}
This work aims to reconstruct ODT images from highly sparsely sampled ODT raw signals to significantly reduce both scanning time and storage requirements. Following the definition and notations in~\cite{li2025sparse}, the fundamental acquisition unit is referred to as the \textbf{raw A-scan}, denoted as $\tilde{\bm{a}}_i$, which encodes depth (A-line) information in the spectral domain. By laterally translating the sensor along the width (B-line), a 2D cross-sectional signal, termed a \textbf{raw B-scan} $\tilde{\bm{B}} = \{\tilde{\bm{a}}_i\}_{i=1}^W$, is obtained, where $W$ represents the number of raw A-scans. 1D inverse fast Fourier transform (IFFT) is applied to each raw A-scan $\tilde{\bm{a}}_i$ to extract the corresponding magnitude $\bm{M} = \{\bm{m}_i\}_{i=1}^W$ and phase $\bm{P} = \{\bm{p}_i\}_{i=1}^W$ responses along the depth in the spatial domain. In the traditional reconstruction pipeline~\cite{zhao2000phase}, the phase differences between adjacent A-lines are utilized to estimate blood flow speed, based on the Doppler frequency shift induced by moving red blood cells. The magnitude response is primarily used for post-processing purposes.

To effectively address the dense scanning issue of the traditional pipeline, we propose a sparse ODT reconstruction pipeline (Figure~\ref{fig:pipeline}). As discussed before, we aggressively sample the raw B-scans with a large stride $\delta$. The magnitude and phase of sparsely sampled A-lines, denoted as $\bm{M}_s = \{\bm{m}_{1+(j-1)\delta}\}_{j=1}^{W'}$ and $\bm{P}_s = \{\bm{p}_{1+(j-1)\delta}\}_{j=1}^{W'}$, are the inputs to the proposed pipeline.

\subsection{Overall Pipeline of ASBA}
The overall pipeline of the proposed flow-aware A-line ROI SSM and B-line phase Attention network (ASBA) for sparse ODT reconstruction is presented in Figure~\ref{fig:pipeline}. Different from the previous work~\cite{li2025sparse} that concatenates the magnitude $\bm{M}_s$ and phase $\bm{P}_s$ as the input to the network, we propose a two-branch pipeline that separately processes $\bm{M}_s$ and $\bm{P}_s$ and focuses on the flow-aware phase modeling. Specifically, $\bm{M}_s$ and $\bm{P}_s$ are first encoded by a convolution layer to deep features individually, processed by a sequence of $n_g$ Magnitude-Phase (\texttt{MP}) Groups, and upscaled by the B-line Shuffle ($\texttt{B-Shuf}$) layer~\cite{shi2016real,li2025sparse} to predict the dense magnitude $\hat{\bm{M}}$ and phase $\hat{\bm{P}}$.  Finally, the features of predicted dense $\hat{\bm{M}}$ and $\hat{\bm{P}}$ are concatenated to predict the final reconstructed B-scan image $\hat{\bm{Y}}$. 

The core structure of the pipeline is the MP Group, consisting of three main components: a Magnitude Module (MagM), a Phase Module (PhaM), and a Fusion block.

Each \textbf{Magnitude Module} (Figure~\ref{fig:block}(a)) comprises $N$ Magnitude Layers, where each layer consists of two Residual Channel Attention Blocks (RCAB)~\cite{zhang2018image}. Given that the magnitude response primarily serves an auxiliary role in ODT reconstruction, we adopt RCAB, a widely used CNN-based module for pixel-wise feature extraction to achieve a balance between effectiveness and computational efficiency. The primary focus of this work is on extracting flow-related information from the phase response.

The \textbf{Fusion Block} (Figure~\ref{fig:block}(b)) is designed to facilitate information exchange between the magnitude and phase branches, thereby enhancing feature extraction in both branches. Specifically, it begins by concatenating feature maps from the two branches, followed by a channel-wise squeeze-expand operation implemented using two convolutional layers to fuse the information effectively and efficiently. The resulting fused feature is then split and integrated back into the respective branches, allowing both branches to benefit from complementary contextual cues.

Each \textbf{Phase Module} (Figure~\ref{fig:block}(c)) consists of $N$ Phase Layers. Because the information distributions differ between A-line and B-line, and only B-line is sparsely sampled, we refer to~\cite{li2025sparse} to model these axes individually. In each layer, an A-line ROI SSM (A-RS) block (Figure~\ref{fig:block}(d), detailed in Sec. 3.3) is first applied to each A-line. This block leverages a modulated Mamba model to selectively extract flow-related features by emphasizing ROI along depth. Subsequently, a B-line Phase Attention (B-PA) block (Figure~\ref{fig:block}(e), detailed in Sec. 3.4) processes each B-line to capture long-range dependencies in the lateral direction, guided by phase differences induced by the Doppler effect. Finally, a feed-forward network (FFN), composed of sequential 1D convolutions along A-line and B-line axes, is employed for feature fusion, following the design in~\cite{li2025sparse}.

\begin{figure*}[!t]
\centering
\includegraphics[width=0.9\linewidth]{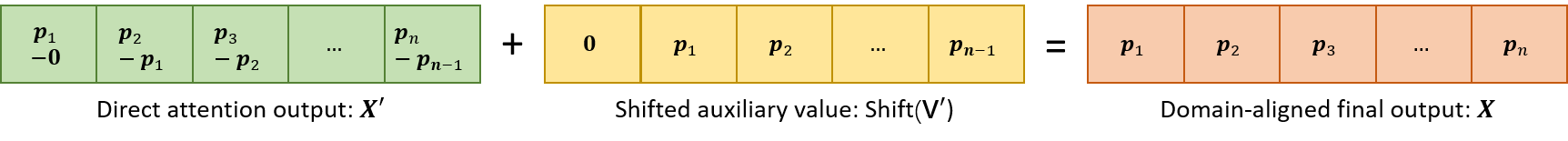}
    \caption{Operations in the B-PA Block to map the phase difference feature back to the phase domain.}
\label{fig:shift}
\end{figure*}

\subsection{A-line ROI State Space Model (A-RS)}
In ODT, the flow signal along an A-line is typically sparse, while background regions occupy a large portion of the depth. As shown in the B-scan image in Figure~\ref{fig:pipeline}, these background regions contribute little to the modeling of blood flow features but often dominate standard modeling approaches. Classical attention-based methods, which rely on computing pairwise similarities across all spatial positions, are not well-suited for such a setting: the abundance of background signals tends to overwhelm the sparse flow signals, reducing both the efficiency and effectiveness of flow feature extraction.

In contrast, state space models (SSMs) offer a more suitable alternative for A-line modeling due to their learned filtering mechanisms and linearly scalable complexity. A notable example is the recently proposed Mamba model~\cite{gu2023mamba}, whose scanning operation is formulated as:
\begin{equation}
    \begin{split}
        h_k = \Bar{\bm{A}} h_{k-1} + \Bar{\bm{B}} x_k, ~~
        y_k = \bm{C}h_k + \bm{D} x_k \\
    \end{split}
\end{equation}
where $\Bar{\bm{A}} = \exp{(\bm{\Delta} \bm{A})}$ is the state matrix, $\Bar{\bm{B}} = (\bm{\Delta} \bm{A})^{-1}(\exp{(\bm{\Delta} \bm{A})} - \bm{I}) \cdot \bm{\Delta} \bm{B}$ is the input matrix, $\bm{C}$ is the output matrix, and $\bm{\Delta}$ is the timescale parameter to discretize $\bm{A}$ and $\bm{B}$. Although Mamba achieves high computational efficiency, its application to A-line modeling in ODT is suboptimal due to the overwhelming influence of non-informative background signals during state propagation.

Addressing this limitation, we propose the A-line ROI State Space Model (A-RS) block (Figure~\ref{fig:block}(d)), which focuses on flow-relevant regions within A-line. To mitigate the influence of non-informative background features, A-RS introduces a masked input strategy, which adaptively suppresses non-informative background signals and noise from the SSM’s input matrix prior to state transformation. This selective modeling enables the system to allocate its representational capacity to flow regions, significantly enhancing the extraction of flow features along the depth axis.

First, we extract a layer-shared feature $\bm{F}$ from the concatenated input magnitude $\bm{M}_s$ and phase $\bm{P}_s$ to identify potential flow regions. The inclusion of magnitude is motivated by its utility in traditional ODT pipelines for suppressing background noise, thereby aiding in distinguishing flow from non-flow areas. Second, for each Phase Layer, a layer-specific convolution followed by \texttt{SiLU}~\cite{elfwing2018sigmoid} is applied to generate an ROI soft mask $\bm{R}$, which adaptively highlights flow regions and suppresses background. Finally, this mask $\bm{R}$ is applied to the Mamba input matrix $\Bar{\bm{B}}$ through element-wise multiplication, enabling flow-aware feature extraction along the depth. The A-line ROI Selective Scan (A-RSS) block of the modulated Mamba can be represented as:
\begin{equation}
\begin{split}
        \bm{F} &= \texttt{Conv} (\bm{M}_s \oplus \bm{P}_s),~
        \bm{R} = \texttt{SiLU} (\texttt{Conv}(\bm{F})) \\
        h_k &= \Bar{\bm{A}} h_{k-1} + (\bm{R} \cdot \Bar{\bm{B}}) x_k, ~
        y_k = \bm{C}h_k + \bm{D} x_k 
\end{split}
\end{equation}

\subsection{B-line Phase Attention (B-PA)}
To capture the long-range flow dependency along the B-line, we introduce the B-line Phase Attention (B-PA) block, as illustrated in Figure~\ref{fig:block}(e). Since the B-line is largely homogeneous and flow signals in raw B-scans often exhibit long-range continuity along the B-line due to dense sampling, the attention mechanism is well-suited for modeling flow dependencies along the B-line~\cite{li2025sparse}. However, directly applying conventional attention modules to phase signals is suboptimal because the characteristics of ODT phase distributions differ significantly from those of other data modalities, such as natural images. 

In ODT, flow-related similarities are not captured by raw phase values themselves, but rather by phase differences between adjacent A-lines, which encode motion-induced changes in signal. Moreover, due to the periodic nature of phase (defined within $(-\pi, \pi]$), flow regions exhibiting similar behavior may yield phase differences with opposite signs, despite representing equivalent physical motion. To ensure the attention mechanism correctly captures such symmetric relationships, we compute the absolute value of the B-line difference (\texttt{B-Diff}) of the query $\bm{Q}$ and key $\bm{K}$ of the attention module as a more robust similarity measure.

The core of the proposed B-PA block (excluding the gate and output modules) can be represented as:
\begin{equation}
\nonumber
    \begin{split}
        \bm{Q}_b, \bm{K}_b &= \texttt{B-Diff}(\bm{Q}), \texttt{B-Diff}(\bm{K})\\
        \bm{Q}_a, \bm{K}_a &= \texttt{DConv}(|\bm{Q}_b|), \texttt{DConv}(|\bm{K}_b|) \\
        \bm{V}_b &= \texttt{B-Diff}(\bm{V}), ~~\bm{V}' = \texttt{Linear}(\bm{V}) \\
        \bm{X}' &= \texttt{B-SA} (\bm{Q}_a, \bm{K}_a, \bm{V}_b), ~~\bm{X} = \bm{X}' + \texttt{Shift}(\bm{V}')
    \end{split} 
\end{equation}
where \texttt{DConv} denotes depth-wise convolution, and \texttt{B-SA} denotes B-line Self-Attention~\cite{li2025sparse}, calculating the 1D attention~\cite{vaswani2017attention} along each B-line. Since the B-line difference operation pads features before B-line subtraction, to map the phase difference feature $\bm{X}'$ back to the phase domain, the auxiliary value $\bm{V}'$ should be shifted accordingly and added to $\bm{X}'$, as shown in Figure~\ref{fig:shift}.

\subsection{Flow-aware Weighted Loss}
Since the majority of an ODT B-scan comprises background regions, which may exhibit noise, applying a uniform mean square error (MSE) loss across all pixels may bias the network toward fitting the background rather than reconstructing flow signals. To address this issue, we propose a flow-aware weighting strategy that assigns higher importance to flow-dominant regions. Specifically, each pixel is weighted based on its groundtruth flow. Given the groundtruth flow value $y_i$ at the $i$-th pixel of a B-scan image, the corresponding weight is calculated as follows:
\begin{equation}
    w_i = y_i^{\alpha} + \beta
\end{equation}
where $\alpha$ controls the balance among different flow values and $\beta$ prevents the background weights from being zero.

Then, we calculate the weighted MSE loss for the prediction of the reconstructed flow $\mathcal{L}_Y$, as well as the magnitude $\mathcal{L}_M$ and phase $\mathcal{L}_P$. For example, $\mathcal{L}_Y$ is calculated by:
\begin{equation}
   \mathcal{L}_{Y} = \frac{\sum{w_i} (\hat{y}_i- y_i)^2}{\sum{w_i}}
\end{equation}

Finally, the total loss is a weighted sum of the three losses with a parameter $\lambda$:
\begin{equation}
    \mathcal{L} = \mathcal{L}_Y + \lambda \cdot (\mathcal{L}_M + \mathcal{L}_P)
\end{equation}


\begin{table*}[!tbp]
\centering
\small
\caption{Comparison of sparse ODT reconstruction results}
\setlength{\tabcolsep}{1mm}
\begin{tabular}{c|l|cccc|cccc}
\hline
\multicolumn{1}{l|}{\multirow{3}{*}{}} & \multirow{3}{*}{Method} & \multicolumn{4}{c|}{MCD-AW}                                                              & \multicolumn{4}{c}{MCD-AN}                                                               \\ \cline{3-10} 
\multicolumn{1}{l|}{}                  &                         & \multicolumn{2}{c|}{B-scan}                           & \multicolumn{2}{c|}{MIP}         & \multicolumn{2}{c|}{B-scan}                           & \multicolumn{2}{c}{MIP}          \\ \cline{3-10} 
\multicolumn{1}{l|}{}                  &                         & PSNR           & \multicolumn{1}{c|}{SSIM}            & PSNR           & SSIM            & PSNR           & \multicolumn{1}{c|}{SSIM}            & PSNR           & SSIM            \\ \hline
\multirow{11}{*}{\rotatebox{90}{$\times8$ Sparsity}}          & Traditional             & 14.32$_{\pm1.35}$          & \multicolumn{1}{c|}{0.2623$_{\pm.0586}$}          & 14.12$_{\pm0.32}$          & 0.2669$_{\pm.0396}$          & 14.37$_{\pm1.53}$          & \multicolumn{1}{c|}{0.2684$_{\pm.0588}$}          & 14.52$_{\pm0.72}$          & 0.2821$_{\pm.0489}$          \\
                                       & SwinIR~\cite{liang2021swinir}                  & 18.81$_{\pm1.17}$          & \multicolumn{1}{c|}{0.4549$_{\pm.0402}$}          & 19.27$_{\pm1.35}$          & 0.4583$_{\pm.0458}$          & 18.57$_{\pm1.35}$          & \multicolumn{1}{c|}{0.4294$_{\pm.0371}$}          & 19.27$_{\pm1.10}$          & 0.4436$_{\pm.0442}$          \\
                                       & HAT~\cite{chen2023activating}                     & 18.71$_{\pm1.26}$          & \multicolumn{1}{c|}{0.4552$_{\pm.0416}$}          & 19.20$_{\pm1.34}$          & 0.4578$_{\pm.0445}$          & 18.71$_{\pm1.31}$          & \multicolumn{1}{c|}{0.4466$_{\pm.0371}$}          & 19.39$_{\pm1.06}$          & 0.4598$_{\pm.0421}$          \\
                                       & SRFormer~\cite{zhou2023srformer}                & 18.70$_{\pm1.31}$          & \multicolumn{1}{c|}{0.4613$_{\pm.0394}$}          & 19.42$_{\pm1.36}$          & 0.4676$_{\pm.0429}$          & 18.78$_{\pm1.36}$          & \multicolumn{1}{c|}{0.4434$_{\pm.0345}$}          & 19.45$_{\pm0.98}$          & 0.4517$_{\pm.0355}$          \\
                                       & DAT~\cite{chen2023dual}                     & 19.23$_{\pm1.33}$          & \multicolumn{1}{c|}{0.5036$_{\pm.0482}$}          & 19.90$_{\pm1.21}$          & 0.5076$_{\pm.0374}$          & 19.04$_{\pm1.26}$          & \multicolumn{1}{c|}{0.4577$_{\pm.0358}$}          & 19.64$_{\pm0.96}$          & 0.4764$_{\pm.0441}$          \\
                                       & CFAT~\cite{ray2024cfat}                    & 18.54$_{\pm1.18}$          & \multicolumn{1}{c|}{0.4287$_{\pm.0356}$}          & 18.89$_{\pm1.17}$          & 0.4237$_{\pm.0381}$          & 18.76$_{\pm1.22}$          & \multicolumn{1}{c|}{0.4507$_{\pm.0383}$}          & 19.38$_{\pm1.07}$          & 0.4648$_{\pm.0421}$          \\
                                       & MambaIR~\cite{guo2024mambair}                 & 18.97$_{\pm1.39}$          & \multicolumn{1}{c|}{0.4811$_{\pm.0480}$}          & 19.67$_{\pm1.31}$          & 0.4948$_{\pm.0437}$          & 19.21$_{\pm1.20}$          & \multicolumn{1}{c|}{0.4826$_{\pm.0415}$}          & 20.15$_{\pm0.95}$          & 0.4997$_{\pm.0358}$          \\
                                       & MambaIRv2~\cite{guo2025mambairv2}               & 18.73$_{\pm1.30}$          & \multicolumn{1}{c|}{0.4526$_{\pm.0453}$}          & 19.18$_{\pm1.34}$          & 0.4523$_{\pm.0546}$          & 18.87$_{\pm1.20}$          & \multicolumn{1}{c|}{0.4532$_{\pm.0380}$}          & 19.52$_{\pm1.04}$          & 0.4646$_{\pm.0417}$          \\
                                       & MaIR~\cite{li2025mair}                    & 19.46$_{\pm1.33}$          & \multicolumn{1}{c|}{0.5129$_{\pm.0528}$}          & 20.10$_{\pm1.30}$          & 0.5235$_{\pm.0389}$          & 19.19$_{\pm1.35}$          & \multicolumn{1}{c|}{0.4872$_{\pm.0446}$}          & 20.15$_{\pm0.85}$          & 0.5091$_{\pm.0313}$          \\
                                       & ASSAN~\cite{li2025sparse}                  & 19.61$_{\pm1.48}$          & \multicolumn{1}{c|}{0.5315$_{\pm.0533}$}          & 20.65$_{\pm1.25}$          & 0.5430$_{\pm.0294}$          & 19.50$_{\pm1.67}$          & \multicolumn{1}{c|}{0.5068$_{\pm.0524}$}          & 20.46$_{\pm1.24}$          & 0.5271$_{\pm.0340}$          \\
                                       & ASBA (Ours)             & \textbf{20.48}$_{\pm1.50}$ & \multicolumn{1}{c|}{\textbf{0.5868}$_{\pm.0559}$} & \textbf{21.65}$_{\pm1.02}$ & \textbf{0.6527}$_{\pm.0169}$ & \textbf{20.38}$_{\pm1.61}$ & \multicolumn{1}{c|}{\textbf{0.5686}$_{\pm.0561}$} & \textbf{21.66}$_{\pm1.05}$ & \textbf{0.6714}$_{\pm.0346}$ \\ \hline
\multirow{11}{*}{\rotatebox{90}{$\times16$ Sparsity}}         & Traditional             & 11.67$_{\pm0.90}$          & \multicolumn{1}{c|}{0.1072$_{\pm.0177}$}          & 7.05$_{\pm0.44}$           & 0.0715$_{\pm.0120}$          & 12.01$_{\pm1.10}$          & \multicolumn{1}{c|}{0.1197$_{\pm.0209}$}          & 7.37$_{\pm0.64}$           & 0.0859$_{\pm.0211}$          \\
                                       & SwinIR~\cite{liang2021swinir}                  & 17.26$_{\pm1.12}$          & \multicolumn{1}{c|}{0.3281$_{\pm.0497}$}          & 17.54$_{\pm1.48}$          & 0.3278$_{\pm.0779}$          & 16.84$_{\pm1.32}$          & \multicolumn{1}{c|}{0.2885$_{\pm.0605}$}          & 17.14$_{\pm0.94}$          & 0.3208$_{\pm.0694}$          \\
                                       & HAT~\cite{chen2023activating}                     & 17.48$_{\pm1.23}$          & \multicolumn{1}{c|}{0.3439$_{\pm.0526}$ }         & 17.66$_{\pm1.35}$          & 0.3329$_{\pm.0715}$          & 17.21$_{\pm1.32}$          & \multicolumn{1}{c|}{0.3255$_{\pm.0583}$}          & 17.60$_{\pm1.01}$          & 0.3550$_{\pm.0726}$          \\
                                       & SRFormer~\cite{zhou2023srformer}                & 17.20$_{\pm1.20}$          & \multicolumn{1}{c|}{0.3150$_{\pm.0547}$ }         & 17.43$_{\pm1.46}$          & 0.3296$_{\pm.0773}$          & 17.10$_{\pm1.34}$          & \multicolumn{1}{c|}{0.3032$_{\pm.0553}$}          & 17.58$_{\pm0.90}$          & 0.3603$_{\pm.0666}$          \\
                                       & DAT~\cite{chen2023dual}                     & 17.52$_{\pm1.27}$          & \multicolumn{1}{c|}{0.3542$_{\pm.0564}$}          & 17.58$_{\pm1.44}$          & 0.3216$_{\pm.0713}$          & 17.44$_{\pm1.29}$          & \multicolumn{1}{c|}{0.3453$_{\pm.0516}$}          & 17.75$_{\pm0.97}$          & 0.3422$_{\pm.0621}$          \\
                                       & CFAT~\cite{ray2024cfat}                    & 17.26$_{\pm1.15}$          & \multicolumn{1}{c|}{0.3353$_{\pm.0530}$}          & 17.51$_{\pm1.42}$          & 0.3265$_{\pm.0759}$          & 17.31$_{\pm1.36}$          & \multicolumn{1}{c|}{0.3210$_{\pm.0525}$}          & 17.65$_{\pm0.99}$          & 0.3545$_{\pm.0725}$          \\
                                       & MambaIR~\cite{guo2024mambair}                 & 17.60$_{\pm1.35}$          & \multicolumn{1}{c|}{0.3614$_{\pm.0613}$}          & 17.70$_{\pm1.47}$          & 0.3236$_{\pm.0667}$          & 17.67$_{\pm1.24}$          & \multicolumn{1}{c|}{0.3579$_{\pm.0515}$}          & 18.10$_{\pm1.02}$          & 0.3577$_{\pm.0623}$          \\
                                       & MambaIRv2~\cite{guo2025mambairv2}               & 17.39$_{\pm1.28}$          & \multicolumn{1}{c|}{0.3473$_{\pm.0560}$}          & 17.55$_{\pm1.33}$          & 0.3241$_{\pm.0695}$          & 16.87$_{\pm1.37}$          & \multicolumn{1}{c|}{0.3010$_{\pm.0584}$}          & 17.38$_{\pm0.93}$          & 0.3476$_{\pm.0712}$          \\
                                       & MaIR~\cite{li2025mair}                    & 17.60$_{\pm1.25}$          & \multicolumn{1}{c|}{0.3697$_{\pm.0545}$}          & 17.77$_{\pm1.43}$          & 0.3297$_{\pm.0640}$          & 17.74$_{\pm1.22}$          & \multicolumn{1}{c|}{0.3714$_{\pm.0497}$}          & 18.47$_{\pm0.89}$          & 0.3649$_{\pm.0522}$          \\
                                       & ASSAN~\cite{li2025sparse}                   & 18.00$_{\pm1.40}$          & \multicolumn{1}{c|}{0.4074$_{\pm.0572}$}          & 18.45$_{\pm1.52}$          & 0.3717$_{\pm.0561}$          & 18.03$_{\pm1.23}$          & \multicolumn{1}{c|}{0.3984$_{\pm.0477}$}          & 18.59$_{\pm0.95}$          & 0.3740$_{\pm.0425}$          \\
                                       & ASBA (Ours)             & \textbf{19.06}$_{\pm1.49}$ & \multicolumn{1}{c|}{\textbf{0.4967}$_{\pm.0535}$} & \textbf{19.88}$_{\pm1.42}$ & \textbf{0.5348}$_{\pm.0404}$ & \textbf{18.95}$_{\pm1.33}$ & \multicolumn{1}{c|}{\textbf{0.4910}$_{\pm.0472}$} & \textbf{20.08}$_{\pm0.74}$ & \textbf{0.5622}$_{\pm.0417}$ \\ \hline
\end{tabular}
\label{table:result}
\end{table*}

\section{Experiments}
\subsection{ODT Dataset of Mouse Brain Cortex}
\label{sec:dataset}
Due to the complexity of acquiring animal blood flow data, most existing studies rely on relatively small datasets~\cite{li2020deep,jiang2020comparative,li2025sparse} (less than 20). In this work, we constructed two datasets from the brain cortex of awake and anesthetized mice, referred to as the Awake Mouse Cortex Dataset (MCD-AW) and the Anesthetized Mouse Cortex Dataset (MCD-AN), respectively. Each dataset contains 20 3D ODT volumes, with 14 used for training and 6 for testing. The training and testing sets are collected from different mice. Each 3D volume contains 500 2D ODT raw B-scans, with each raw B-scan having a resolution of $256 \times 12600$. The groundtruths are obtained by applying the traditional pipeline to the densely sampled raw signals. Since ODT volumes are typically visualized using en-face Maximum Intensity Projection (MIP), we assess reconstruction quality using PSNR and SSIM on both reconstructed B-scans and MIP images. 
Moreover, because the B-scans are predominantly composed of dark background regions that are less relevant to flow analysis, we restrict evaluation metrics on B-scans to pixels whose groundtruth values exceed 5\% of the maximum possible value (65535 for uint16). 

\subsection{Implementation Details}
\label{sec:details}
We carry out experiments on highly sparsely sampled ODT raw B-scans at $\times 8$ and $\times 16$ sparsity. The implemented ASBA consists of 4 MP Groups. Each magnitude and phase module comprises 6 layers. The deep embedding channels of both magnitude and phase branches are 60. The input patch is of $256\times64$. The parameters are set as $\alpha=0.5$, $\beta=0.1$, and $\lambda=0.5$. The model is trained in batches of 8 for 100K iterations, with Adam~\cite{kingma2014adam} optimizer ($\beta_1=0.9$ and $\beta_2=0.999$) and CosineAnnealingLR~\cite{loshchilov2016sgdr} (initial and minimum learning rates are $2e-4$ and $1e-6$, respectively).

We compare ASBA with the classical ODT reconstruction and the recent ASSAN~\cite{li2025sparse}. In addition, we include state-of-the-art single image super-resolution (SISR) methods—SwinIR~\cite{liang2021swinir}, HAT~\cite{chen2023activating}, SRFormer~\cite{zhou2023srformer}, DAT~\cite{chen2023dual}, CFAT~\cite{ray2024cfat}, MambaIR~\cite{guo2024mambair}, MambaIRv2~\cite{guo2025mambairv2} and MaIR~\cite{li2025mair}—as they also address an upscaling problem. These SISR methods are adapted similarly to ASSAN, where the sparsely sampled magnitude and phase signals are concatenated to form the input patch. For fair comparison, the embedding dimension of all compared methods is set to 120, matching the combined channel size of our magnitude and phase branches. The other network parameters and training configurations are kept consistent with ours.

\subsection{Main Results}
\begin{table}[!tbp]
\centering
\small
\caption{Comparison of vessel segmentation results on the MIP images under $\times16$ sparsity}
\begin{tabular}{l|cc|cc}
\hline
\multirow{2}{*}{Method} & \multicolumn{2}{c|}{MCD-AW}     & \multicolumn{2}{c}{MCD-AN}      \\ \cline{2-5} 
                        & DICE           & DICE$_s$        & DICE           & DICE$_s$        \\ \hline
Traditional             & 44.93          & 38.94          & 47.18          & 39.94          \\
SwinIR~\cite{liang2021swinir}                  & 58.83          & 47.86          & 57.25          & 45.67          \\
HAT~\cite{chen2023activating}                     & 59.56          & 48.18          & 60.02          & 50.84          \\
SRFormer~\cite{zhou2023srformer}                & 58.40          & 47.01          & 59.33          & 49.71          \\
DAT~\cite{chen2023dual}                     & 60.90          & 51.09          & 62.91          & 54.62          \\
CFAT~\cite{ray2024cfat}                    & 59.28          & 48.20          & 61.21          & 51.74          \\
MambaIR~\cite{guo2024mambair}                 & 61.99          & 52.64          & 62.90          & 54.82          \\
MambaIRv2~\cite{guo2025mambairv2}               & 59.66          & 49.31          & 58.41          & 49.05          \\
MaIR~\cite{li2025mair}                    & 61.39          & 51.41          & 64.63          & 56.44          \\
ASSAN~\cite{li2025sparse}                   & 66.00          & 58.07          & 66.65          & 59.09          \\
ASBA (Ours)                    & \textbf{70.28} & \textbf{64.81} & \textbf{71.93} & \textbf{66.17} \\ \hline
\end{tabular}
\label{table:seg}
\end{table}

\textbf{Quantitative Results}. 
Table~\ref{table:result} compares ASBA against traditional ODT reconstruction, SOTA SISR methods, and the recent sparse ODT reconstruction approach under $\times 8$ and $\times 16$ sparsity. Notably, under the most challenging $\times 16$ sparsity, ASBA achieves, on average across both datasets, a gain of around 1.0 dB in B-scan PSNR and 1.5 dB in MIP PSNR compared to SOTA methods. In addition, ASBA consistently improves SSIM, with an average gain of around 0.09 on B-scan images and 0.18 on MIP images. All results demonstrate the effectiveness of the proposed ASBA.

Table~\ref{table:seg} presents vessel segmentation results on MIP under $\times16$ sparsity to evaluate clinical performance. We report DICE scores over the entire image and within a $256\times256$ ROI of each MIP, which primarily contains small vessels, denoted as DICE and DICE$_s$, respectively. Because manual vessel annotation is highly labor-intensive, we employ a UNet~\cite{ronneberger2015u} pretrained on the OCTA500 dataset~\cite{li2024octa} to generate vessel masks for both the groundtruth and sparsely reconstructed images. The results indicate that ASBA achieves more accurate reconstruction compared to other methods, especially the small vessel flows.

\begin{figure*}[!t]
\centering
\includegraphics[width=0.9\linewidth,height=0.75\linewidth]{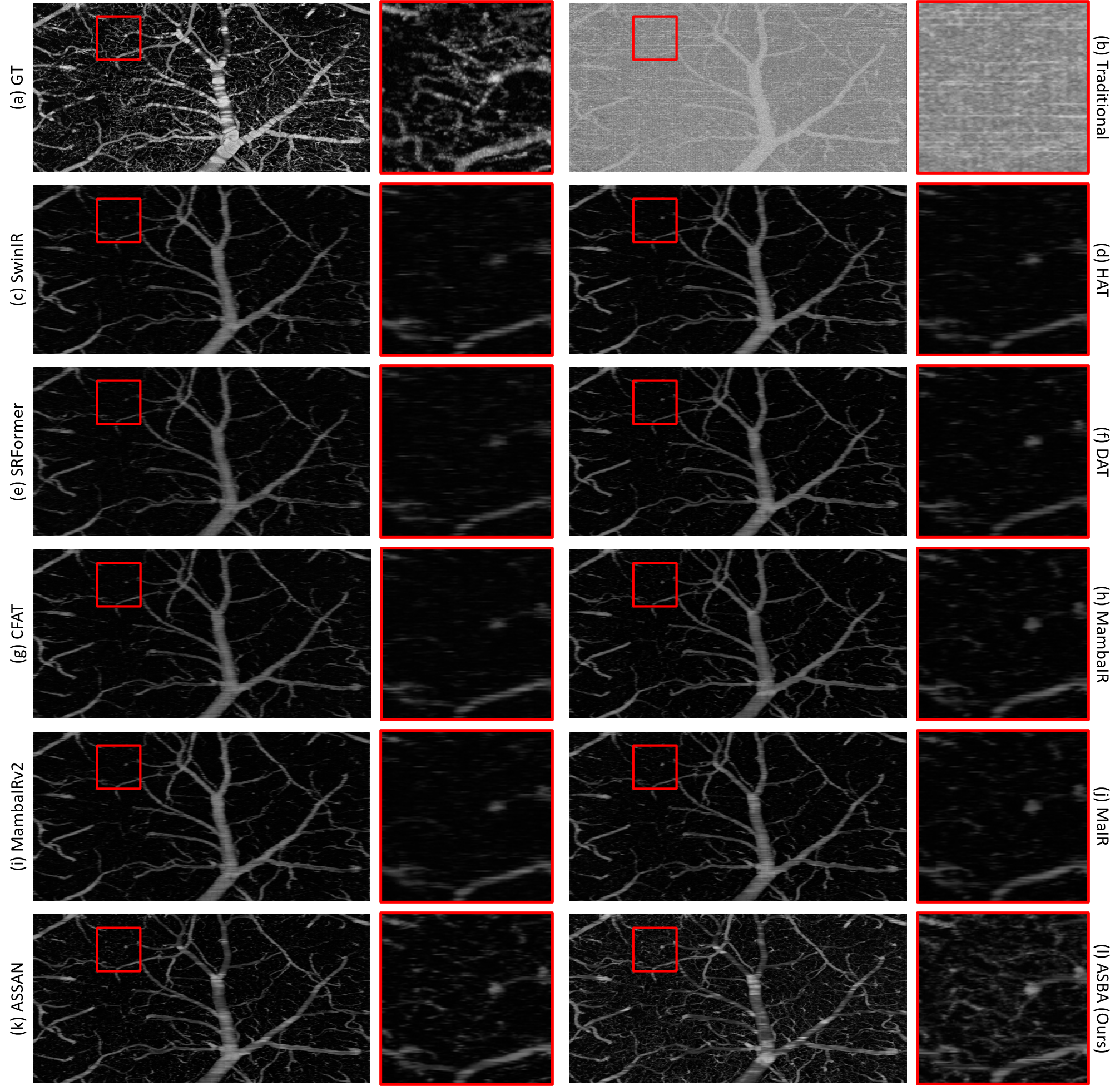}
    \caption{Qualitative results of $\times 16$ sparsity on MCD-AW dataset. Regions of interest are zoomed in.}
\label{fig:result}
\end{figure*}

\begin{table}[!tbp]
\centering
\small
\caption{Ablation Study on A-line ROI SSM (A-RS), B-line Phase Attention (B-PA), flow-aware loss (Loss), and other linear complexity methods for A-line modeling (A-line)}
\begin{tabular}{c|l|cc|cc}
\hline
\multirow{2}{*}{}                          & \multirow{2}{*}{Method} & \multicolumn{2}{c|}{B-scan}      & \multicolumn{2}{c}{MIP}          \\ \cline{3-6} 
                                           &                         & PSNR           & SSIM            & PSNR           & SSIM            \\ \hline
\multirow{4}{*}{\rotatebox{90}{A-RS}}          & w/o ROI                 & 18.89          & 0.4891          & 19.61          & 0.5240          \\
                                           & w/o $\bm{M}_s$                 & 18.94          & 0.4915          & 19.70          & 0.5321          \\
                                           & ROI on $\bm{C}$                & 18.91          & 0.4869          & 19.70          & 0.5256          \\
                                           & ROI on $\bm{B}$                & \textbf{19.06} & \textbf{0.4967} & \textbf{19.88} & \textbf{0.5348} \\ \hline
\multirow{4}{*}{\rotatebox{90}{B-PA}}          & B-GA                    & 18.90          & 0.4889          & 19.70          & 0.5278          \\
                                           & w/o ABS                 & 18.96          & 0.4927          & 19.67          & 0.5310          \\
                                           & w/o Shift               & 18.76          & 0.4920          & 19.56          & 0.5263          \\
                                           & B-PA                    & \textbf{19.06} & \textbf{0.4967} & \textbf{19.88} & \textbf{0.5348} \\ \hline
\multirow{4}{*}{\rotatebox{90}{Loss}} & w/o $w_i$                  & 18.15          & 0.4132          & 18.59          & 0.3886          \\
                                           & $\alpha=2$                 & 18.09          & 0.4477          & 19.24          & 0.4651          \\
                                           & $\alpha=1$                 & 18.81          & 0.4779          & 19.64          & 0.5099          \\
                                           & $\alpha=0.5$               & \textbf{19.06} & \textbf{0.4967} & \textbf{19.88} & \textbf{0.5348} \\ \hline
\multirow{3}{*}{\rotatebox{90}{A-line}} & MILA~\cite{han2024demystify}                    & 18.64          & 0.4787          & 19.35          & 0.5025          \\
& VRWKV~\cite{duanvision}                   & 18.49          & 0.4715          & 19.17          & 0.5125          \\
& A-RS (Ours)                   & \textbf{19.06} & \textbf{0.4967} & \textbf{19.88} & \textbf{0.5348} \\ \hline
\end{tabular}
\label{table:ablation}
\end{table}

\textbf{Qualitative Results}.
Figure~\ref{fig:result} provides a visual comparison of MIP reconstructions under $\times 16$ sparsity on MCD-AW. This high sparsity level poses significant challenges, particularly for recovering slow flows in small vascular structures. Traditional reconstruction methods fail to recover such flows due to the wide gaps between sampled raw A-scans. Similarly, SOTA SISR and sparse ODT reconstruction methods fail to reconstruct flow in small vessels. In contrast, our ASBA successfully recovers these delicate flow structures, demonstrating its superior capability in handling highly sparse inputs and preserving fine vascular details. More results are presented in the supplementary.

\subsection{Ablation Study}
All ablation studies are conducted on MCD-AW at $\times 16$ sparsity. The results are all presented in Table~\ref{table:ablation}.

\textbf{A-line ROI SSM}. In the absence of ROI guidance, the classical Mamba module is less effective at modeling flow information, as the ODT B-scan is largely composed of background regions that contribute minimally to flow reconstruction. In contrast, incorporating guidance from $\bm{M}_s$ enhances performance, since magnitude data is effective for identifying background regions. We also evaluate an alternative strategy that applies the ROI to the output matrix $\bm{C}$ instead of the input $\bm{B}$ of Mamba. However, this approach yields inferior results, as the state-space transformation remains influenced by the background, highlighting the importance of modulating the input matrix via ROI guidance.

\textbf{B-line Phase Attention}. First, we compare B-PA with the B-line Gated Attention (B-GA)~\cite{li2025sparse}, which directly applies 1D attention using a gated mechanism. The results show that incorporating B-line phase differences effectively enhances the modeling of long-range flow dependencies along the B-line. Second, computing the absolute value of the B-line phase differences for $\bm{Q}$ and $\bm{K}$ improves similarity matching, as the periodic nature of the phase implies that the sign of the difference does not reliably indicate feature similarity. Finally, omitting the shift operation on the auxiliary value feature leads to a performance drop, indicating that proper alignment is essential for accurately mapping phase difference features back to the phase domain.

\textbf{Flow-aware Weighted Loss}. Removing the weight term $w_i$ leads to a performance drop, as the background noise dominating the B-scans distracts the network from effectively modeling flow features. This issue is exacerbated by the raw magnitude and phase supervision signals, which are not post-processed and contain significant background noise. Hence, a flow-aware loss function is crucial to make the network focus on flow modeling. We also examine the effect of the exponent $\alpha$, which controls the concavity of $w_i$. Compared to an aggressive weighting strategy that overly emphasizes high-flow regions ($\alpha=2$), a more balanced setting ($\alpha=0.5$) yields better performance.

\textbf{A-line Modeling with Linear Complexity}. An A-line in ODT is typically long (e.g., 256 pixels), making classical self-attention infeasible due to quadratic complexity and poor focus on sparse flow regions. To validate the benefit of our proposed A-RS module for A-line flow modeling, we replace it with two recent linear-complexity methods: MILA~\cite{han2024demystify} (linear attention) and VRWKV~\cite{duanvision} (attention-free). While MILA reduces computational cost by computing similarity in a projected space, it remains susceptible to background interference. VRWKV’s spatial decay helps limit distant dependencies but uses a single learned decay embedding, lacking spatial adaptivity. In contrast, A-RS explicitly filters background features through ROI-enhanced modulation, enabling adaptive focus on flow-relevant regions and yielding superior A-line modeling performance.

\textbf{Other Ablation Studies}. We further conducted experiments on other factors of the proposed ASBA. The results show that the two-branch pipeline configuration is reasonable, and ASBA is consistently effective on smaller sparsity settings, etc. Details are presented in the supplementary.

\begin{table}[!tbp]
\centering
\small
\caption{Comparison of computational complexity}
\begin{tabular}{@{\hspace{.51mm}}l@{\hspace{.51mm}} |c|c|c|c| @{\hspace{.951mm}}c@{\hspace{.51mm}}}
\hline
Method & \#param & FLOPs  & Runtime & PSNR  & SSIM   \\ \hline
MambaIR   & 6.73M   & 163.0G & 90.6ms  & 18.97 & 0.4811  \\
MambaIR$_{\rm v2}$  & 7.97M   & 166.3G & 161.2ms  & 18.73 & 0.4526 \\
MaIR   & 6.13M   & 152.0G & 101.4ms & 19.46 & 0.5129 \\
ASSAN  & 7.56M   & 158.8G & 98.1ms  & 19.61 & 0.5315 \\
ASBA   & 7.07M   & 148.0G & 73.9ms  & 20.48 & 0.5868 \\ \hline
\end{tabular}
\label{table:complexity}
\end{table}

\subsection{Computational Complexity Analysis}
Table~\ref{table:complexity} compares the computational complexity and B-scan reconstruction performance on the MCD-AW dataset under $\times 8$ sparsity. The inference runtime is tested for a $256\times64$ input patch on an RTX A6000 GPU. Compared with other SSM-based methods~\cite{guo2024mambair,guo2025mambairv2,li2025mair,li2025sparse}, the proposed ASBA model achieves the best reconstruction performance while requiring the lowest number of FLOPs and inference time. These results demonstrate that ASBA achieves superior performance on both effectiveness and computational efficiency. More results are presented in supplementary.

\section{Conclusion}
We propose ASBA (A-line State space and B-line Attention), a flow-aware network to reconstruct ODT images from highly sparsely sampled raw A-scans, hence reducing scanning time and storage requirement significantly. Specifically, we develop a two-branch architecture to process magnitude and phase separately, an A-line ROI SSM block to guide attention to flow regions, a B-line Phase Attention block for phase difference, and a flow-aware loss to adaptively emphasize flow regions during training. Extensive experiments validate the effectiveness of ASBA.

{
    \small
    \bibliographystyle{ieeenat_fullname}
    \bibliography{ref}
}

\clearpage
\setcounter{page}{1}
\maketitlesupplementary

\section{Additional Ablation Study}
The ablation studies on the two-branch pipeline and the bias term of the loss function are conducted on the MCD-AW dataset under $\times 16$ sparsity. 
\subsection{Ablation on Two-Branch Pipeline}
Table~\ref{table:pipeline} compares various configurations of the proposed two-branch pipeline. The results highlight the critical role of the fusion block in facilitating information exchange between the magnitude and phase branches. Additionally, removing the supervision from the raw magnitude and phase signals ($\lambda=0$) leads to a drop in reconstruction accuracy, indicating that proper guidance from these signals is essential for the training of the two-branch network.

\subsection{Ablation on Bias Term of Loss Function}
Table~\ref{table:beta} compares different bias terms of the flow-aware weighted loss function. A small bias term $\beta$ generally leads to good reconstruction performance. However, when $\beta$ becomes too large (e.g., $\beta=1$), the background regions are assigned relatively higher weights, which negatively affects the overall performance.

\subsection{Ablation on Sparsity}
\begin{figure*}[!t]
\centering
\includegraphics[width=0.85\linewidth]{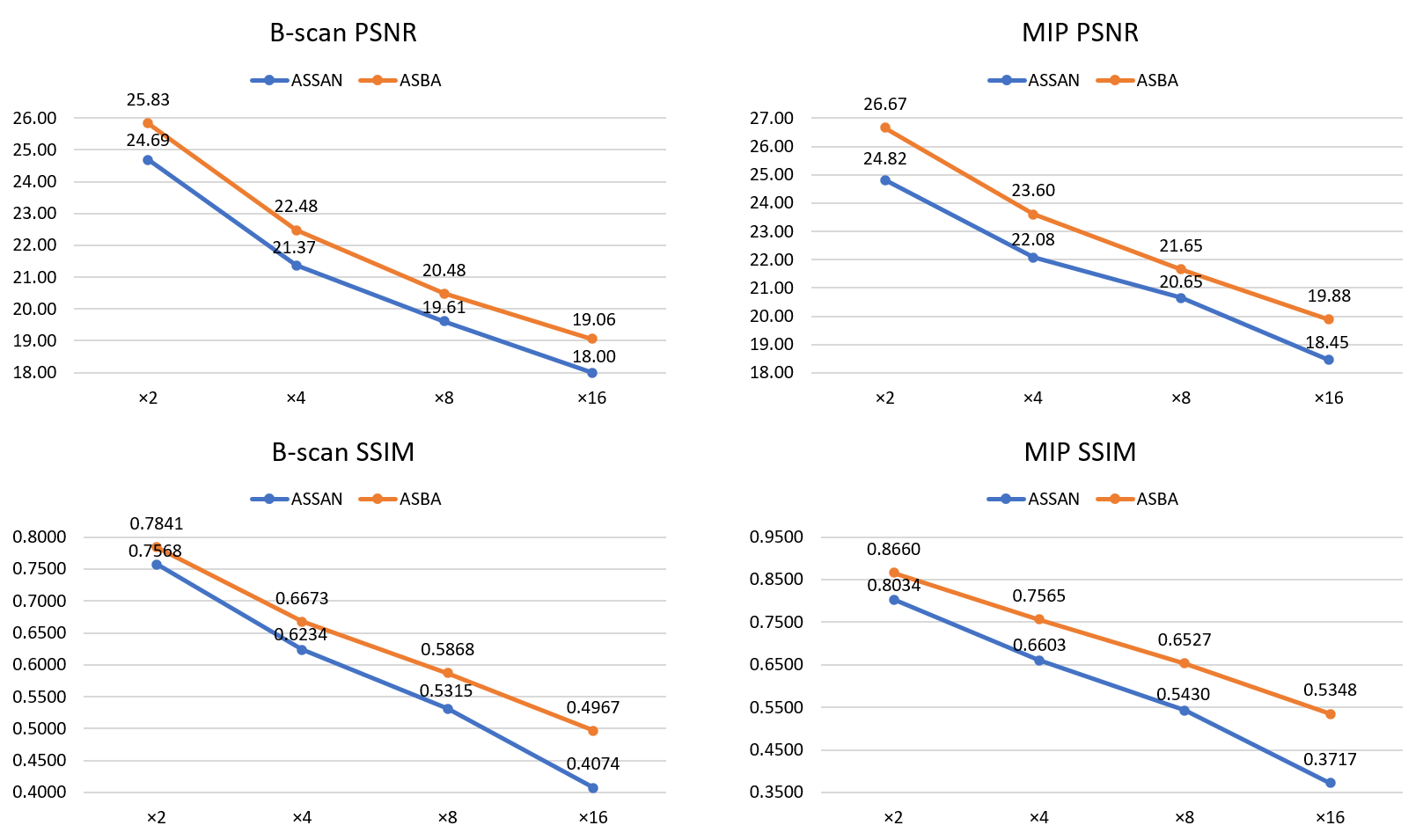}
    \caption{Reconstruction results of various sparsity levels on the MCD-AW dataset.}
\label{fig:sparsity}
\end{figure*}

Figure~\ref{fig:sparsity} compares the performance of the SOTA sparse ODT reconstruction method ASSAN~\cite{li2025sparse} and the proposed ASBA across various sparsity levels on the MCD-AW dataset. The results demonstrate that ASBA consistently outperforms ASSAN under both conservative ($\times 2$, $\times 4$) and aggressive ($\times 8$, $\times 16$) sparsity settings. Notably, for SSIM results, the performance gap between ASBA and ASSAN becomes larger as sparsity increases. These results show the better robustness of ASBA.

\subsection{Ablation on Cross-Dataset Validation}
To further evaluate generalization capabilities of different models, Table~\ref{table:cross} reports both self-dataset validation results (training and testing on the same MCD dataset) and cross-dataset validation results (training on MCD-AW and testing on MCD-AN, or vice versa), denoted as ASBA (Self) and ASBA (Cross). Results are grouped by test sets. For comparison, we also include the self-dataset performance of MaIR~\cite{li2025mair} and ASSAN~\cite{li2025sparse}. ASBA (Cross) continues to substantially outperform MaIR and ASSAN, and ASBA (Cross) trained on MCD-AW achieves performance closely matching ASBA (Self) on the MCD-AN test set. These findings demonstrate the strong cross-dataset generalization ability of the proposed ASBA method.

\section{Computational Complexity Analysis}
Table~\ref{table:complexity} extends \textit{Table 4 of the main paper} and compares the computational complexity and reconstruction performance of both transformer-based methods~\cite{liang2021swinir,chen2023activating,zhou2023srformer,chen2023dual,ray2024cfat} and SSM-based approaches~\cite{guo2024mambair,guo2025mambairv2,li2025mair,li2025sparse} on the MCD-AW dataset under $\times 8$ sparsity. Compared with transformer-based methods, ASBA yields substantial gains in reconstruction accuracy while maintaining competitive computational efficiency. Relative to other SSM-based approaches, ASBA achieves the best FLOPs, the lowest inference time, and the highest reconstruction accuracy. Overall, these results show that ASBA provides a strong balance between effectiveness and computational efficiency.

\begin{table}[!tbp]
\centering
\small
\caption{Comparison of configurations of the two-branch pipeline}
\begin{tabular}{l|cc|cc}
\hline
\multirow{2}{*}{Method} & \multicolumn{2}{c|}{B-scan}      & \multicolumn{2}{c}{MIP}          \\ \cline{2-5} 
                        & PSNR           & SSIM            & PSNR           & SSIM            \\ \hline
w/o fusion              & 18.69          & 0.4805          & 19.48          & 0.5173          \\
$\lambda=0$             & 18.13          & 0.3998          & 18.62          & 0.3989          \\
$\lambda=1$             & 18.78          & 0.4853          & 19.56          & 0.5280         \\
$\lambda=0.5$           & \textbf{19.06} & \textbf{0.4967} & \textbf{19.88} & \textbf{0.5348} \\ \hline
\end{tabular}
\label{table:pipeline}
\end{table}

\begin{table}[!tbp]
\centering
\small
\caption{Comparison of the bias term of the flow-aware weighted loss}
\begin{tabular}{l|cc|cc}
\hline
\multirow{2}{*}{Method} & \multicolumn{2}{c|}{B-scan}      & \multicolumn{2}{c}{MIP}          \\ \cline{2-5} 
                        & PSNR           & SSIM            & PSNR           & SSIM            \\ \hline
$\beta=0$                  & \textbf{19.28} & 0.5090          & 19.67          & 0.4772          \\
$\beta=0.01$               & 19.11          & \textbf{0.5096} & 19.76          & 0.5005          \\
$\beta=0.1$                & 19.06          & 0.4967          & \textbf{19.88} & \textbf{0.5348} \\
$\beta=1$                  & 18.50          & 0.4413          & 18.98          & 0.4481          \\ \hline
\end{tabular}
\label{table:beta}
\end{table}

\begin{table*}[!tbp]
\centering
\small
\caption{Comparison of self and cross-dataset validation}
\begin{tabular}{l|cccc|cccc}
\hline
\multirow{3}{*}{Method} & \multicolumn{4}{c|}{MCD-AW}                                                              & \multicolumn{4}{c}{MCD-AN}                                                               \\ \cline{2-9} 
                        & \multicolumn{2}{c|}{B-scan}                           & \multicolumn{2}{c|}{MIP}         & \multicolumn{2}{c|}{B-scan}                           & \multicolumn{2}{c}{MIP}          \\ \cline{2-9} 
                        & PSNR           & \multicolumn{1}{c|}{SSIM}            & PSNR           & SSIM            & PSNR           & \multicolumn{1}{c|}{SSIM}            & PSNR           & SSIM            \\ \hline
MaIR~\cite{li2025mair}                    & 17.60          & \multicolumn{1}{c|}{0.3697}          & 17.77          & 0.3297          & 17.74          & \multicolumn{1}{c|}{0.3714}          & 18.47          & 0.3649          \\
ASSAN~\cite{li2025sparse}                   & 18.00          & \multicolumn{1}{c|}{0.4074}          & 18.45          & 0.3717          & 18.03          & \multicolumn{1}{c|}{0.3984}          & 18.59          & 0.3740          \\
ASBA (Self)             & \textbf{19.06} & \multicolumn{1}{c|}{\textbf{0.4967}} & \textbf{19.88} & \textbf{0.5348} & 18.95          & \multicolumn{1}{c|}{0.4910}          & 20.08          & \textbf{0.5622} \\
ASBA (Cross)            & 18.75          & \multicolumn{1}{c|}{0.4788}          & 19.35          & 0.5143          & \textbf{19.01} & \multicolumn{1}{c|}{\textbf{0.4946}} & \textbf{20.18} & 0.5563          \\ \hline
\end{tabular}
\label{table:cross}
\end{table*}

\begin{table*}[!tbp]
\centering
\small
\makeatletter\def\@captype{table}
\caption{Comparison of computational complexity}
\begin{tabular}{l|c|c|c|cc|cc}
\hline
\multirow{2}{*}{Method} & \multirow{2}{*}{\#param} & \multirow{2}{*}{FLOPs} & \multirow{2}{*}{Runtime} & \multicolumn{2}{c|}{B-scan} & \multicolumn{2}{c}{MIP} \\ \cline{5-8} 
                        &                          &                        &                          & PSNR         & SSIM         & PSNR       & SSIM       \\ \hline
Traditional             & -                        & -                      & -                        & 14.32        & 0.2623       & 14.12      & 0.2669     \\
SwinIR~\cite{liang2021swinir}                  & 3.77M                    & 72.1G                  & 38.4ms                   & 18.81        & 0.4549       & 19.27      & 0.4583     \\
HAT~\cite{chen2023activating}                     & 6.48M                    & 141.0G                 & 92.8ms                   & 18.71        & 0.4552       & 19.20      & 0.4578     \\
SRFormer~\cite{zhou2023srformer}                & 3.44M                    & 85.3G                  & 66.3ms                   & 18.70        & 0.4613       & 19.42      & 0.4676     \\
DAT~\cite{chen2023dual}                     & 3.57M                    & 74.9G                  & 125.6ms                  & 19.23        & 0.5036       & 19.90      & 0.5076     \\
CFAT~\cite{ray2024cfat}                    & 6.48M                    & 141.0G                 & 124.8ms                  & 18.54        & 0.4287       & 18.89      & 0.4237     \\
MambaIR~\cite{guo2024mambair}                 & 6.73M                    & 163.0G                 & 90.6ms                   & 18.97        & 0.4811       & 19.67      & 0.4948     \\
MambaIRv2~\cite{guo2025mambairv2}               & 7.97M                    & 166.3G                 & 161.2ms                  & 18.73        & 0.4526       & 19.18      & 0.4523     \\
MaIR~\cite{li2025mair}                    & 6.13M                    & 152.0G                 & 101.4ms                  & 19.46        & 0.5129       & 20.10      & 0.5235     \\
ASSAN~\cite{li2025sparse}                   & 7.56M                    & 158.8G                 & 98.1ms                   & 19.61        & 0.5315       & 20.65      & 0.5430     \\
ASBA (Ours)             & 7.07M                    & 148.0G                 & 73.9ms                   & 20.48        & 0.5868       & 21.65      & 0.6527     \\ \hline
\end{tabular}
\label{table:complexity}
\end{table*}

\section{Additional Qualitative Results}
\begin{figure*}[!t]
\centering
\includegraphics[width=\linewidth]{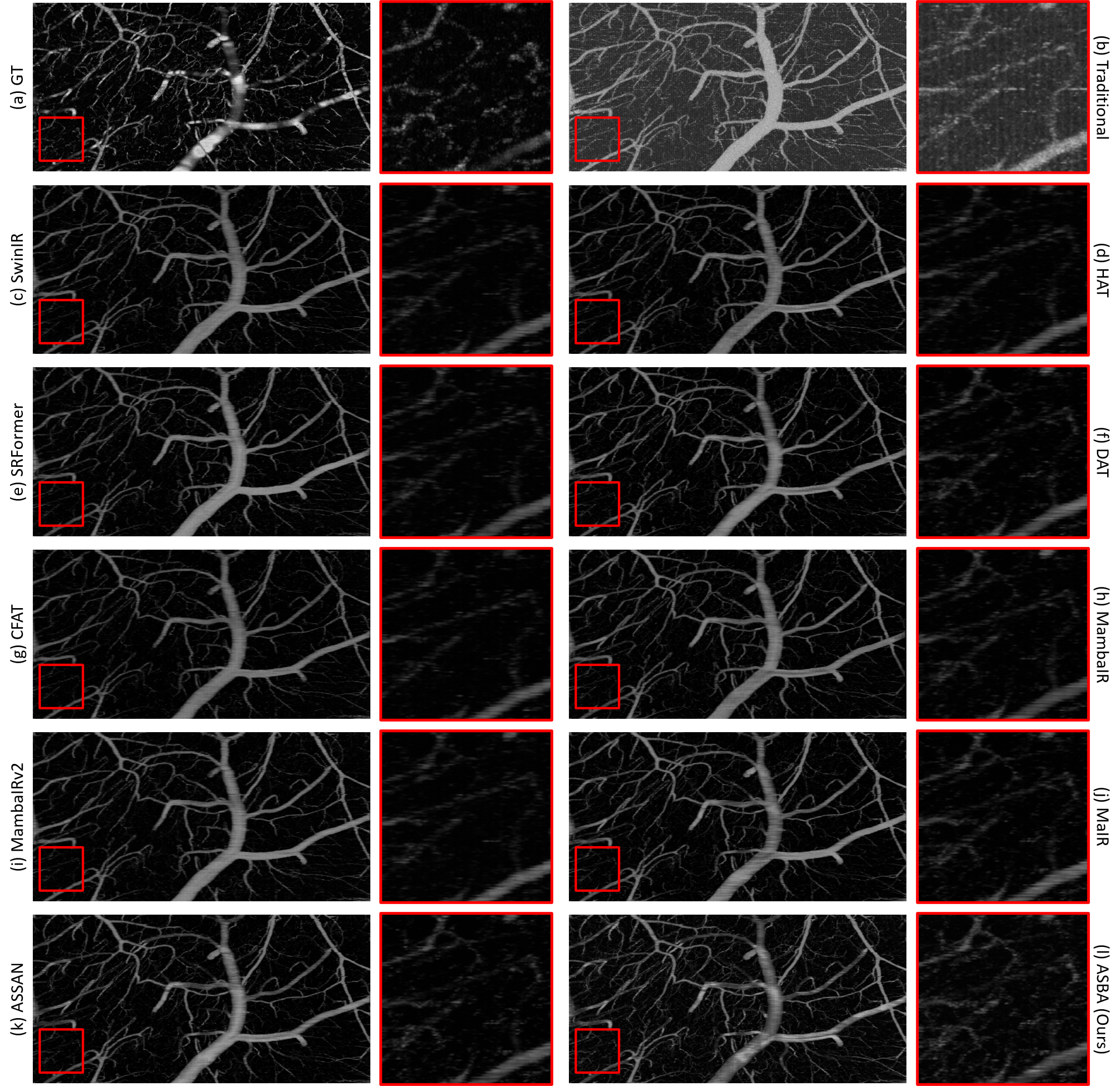}
    \caption{Qualitative results of $\times 8$ sparsity on MCD-AW dataset. Regions of interest are zoomed in.}
\label{fig:ds8aw}
\end{figure*}

\begin{figure*}[!t]
\centering
\includegraphics[width=\linewidth]{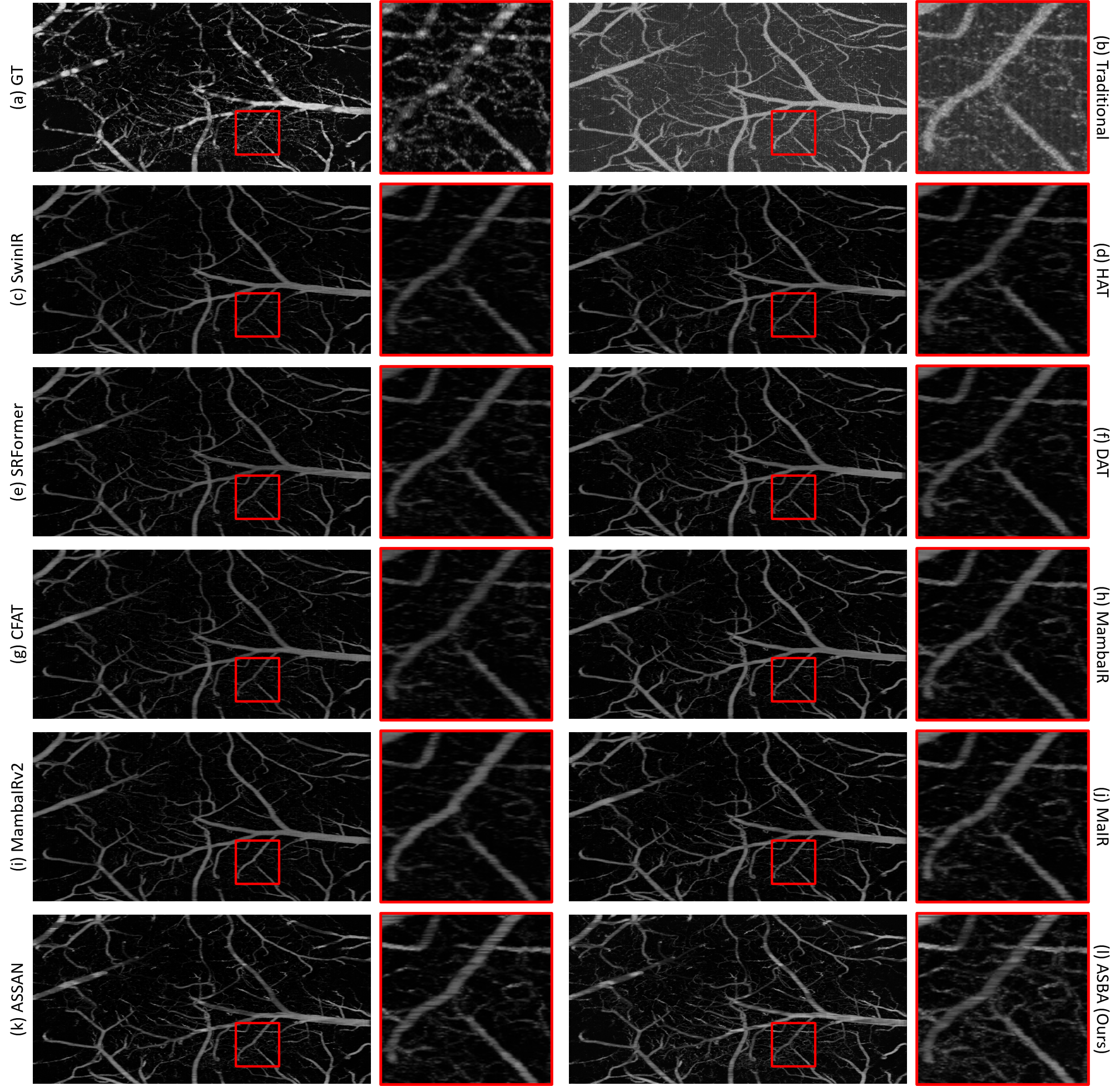}
    \caption{Qualitative results of $\times 8$ sparsity on MCD-AN dataset. Regions of interest are zoomed in.}
\label{fig:ds8an}
\end{figure*}

\begin{figure*}[!t]
\centering
\includegraphics[width=\linewidth]{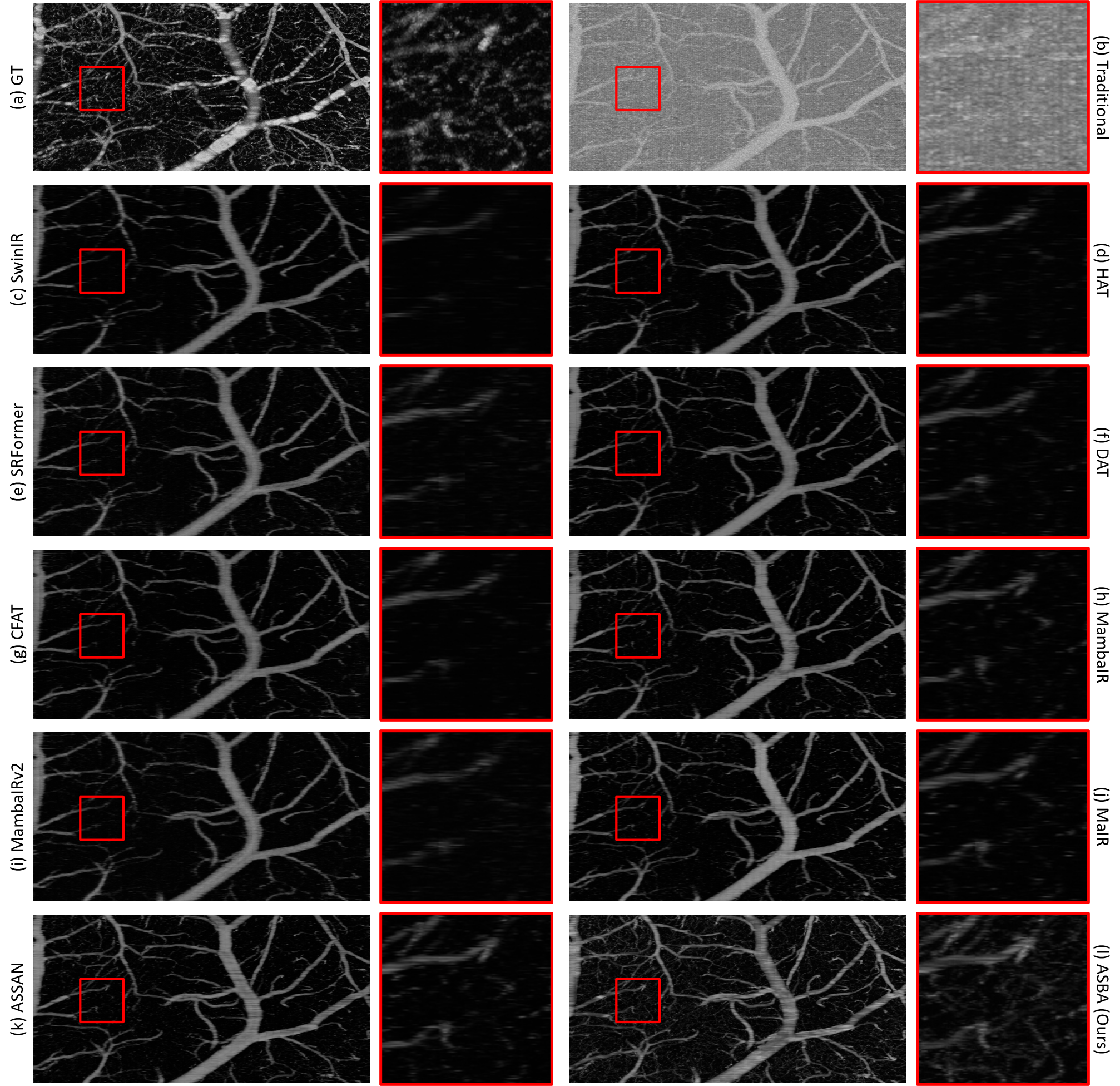}
    \caption{Qualitative results of $\times 16$ sparsity on MCD-AN dataset. Regions of interest are zoomed in.}
\label{fig:ds16an}
\end{figure*}

\begin{figure*}[!t]
\centering
\includegraphics[width=\linewidth]{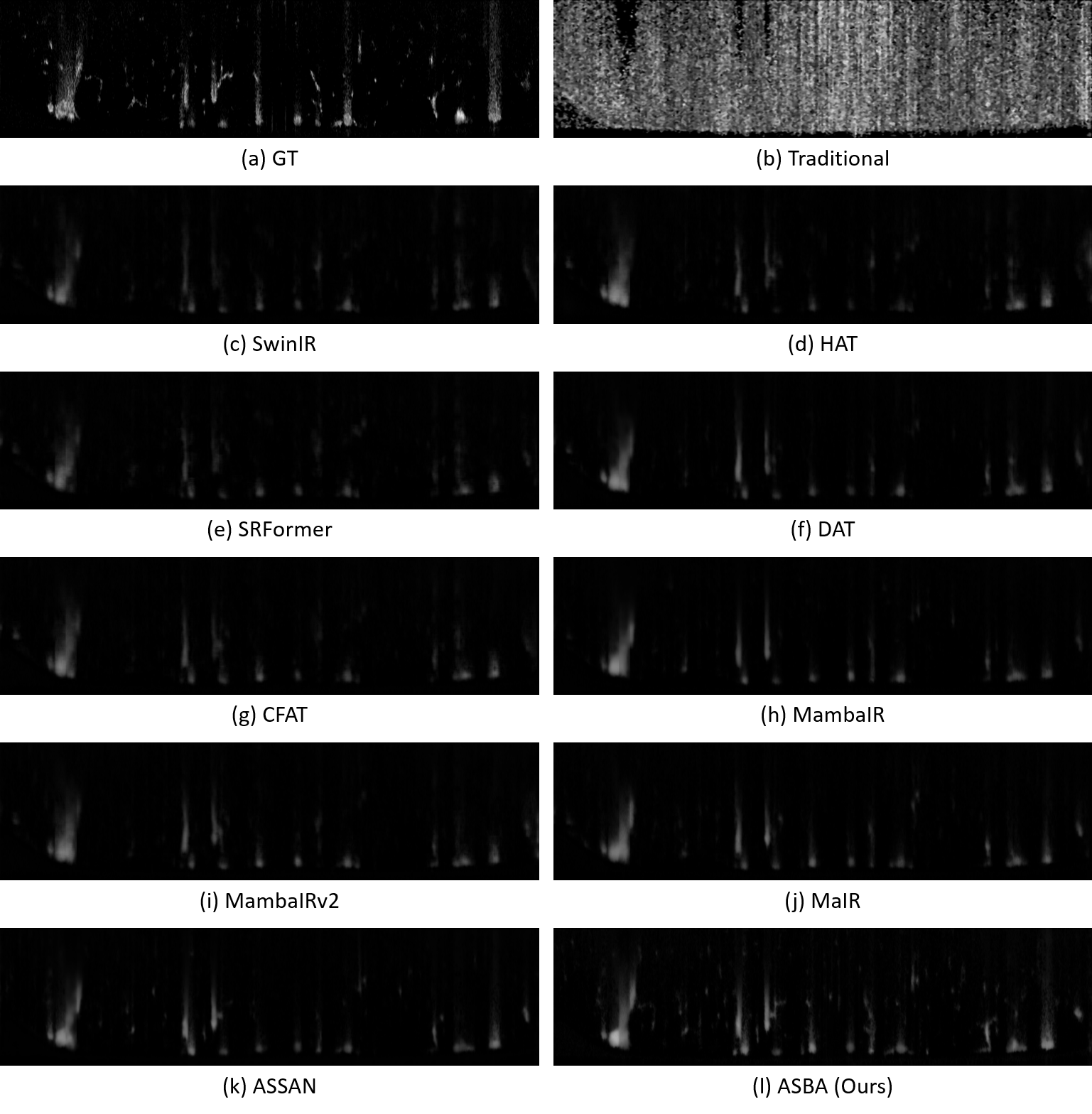}
    \caption{Qualitative results of reconstructed B-scan under $\times 16$ sparsity on MCD-AW dataset.}
\label{fig:ds16bscan-aw}
\end{figure*}

\begin{figure*}[!t]
\centering
\includegraphics[width=\linewidth]{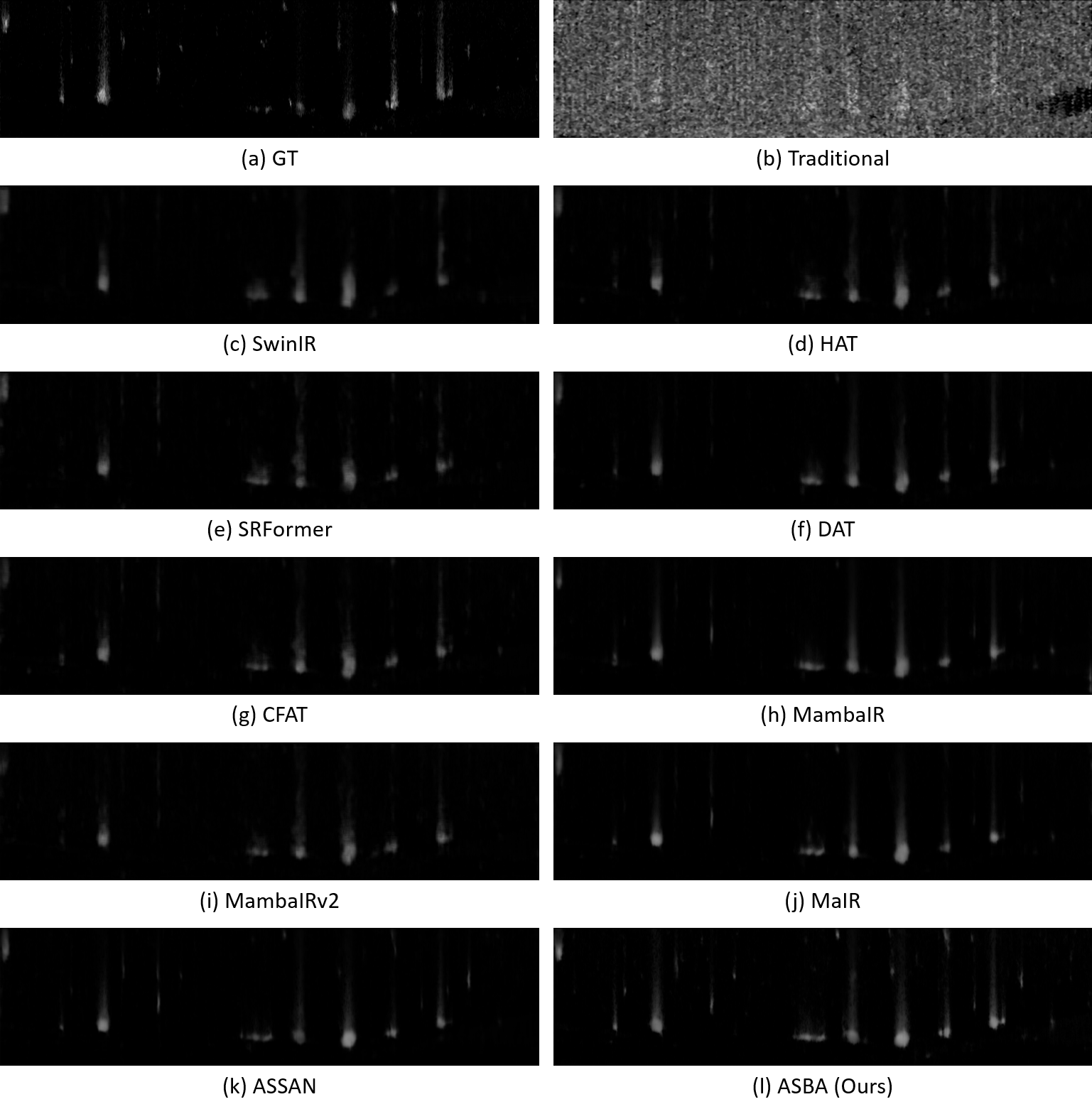}
    \caption{Qualitative results of reconstructed B-scan under $\times 16$ sparsity on MCD-AN dataset.}
\label{fig:ds16bscan-an}
\end{figure*}

Figures~\ref{fig:ds8aw},~\ref{fig:ds8an}, and~\ref{fig:ds16an} present the reconstructed MIP visualizations under $\times 8$ sparsity on the MCD-AW dataset, $\times 8$ sparsity on the MCD-AN dataset, and $\times 16$ sparsity on the MCD-AN dataset respectively for (a) Groundtruth, (b) Traditional Pipeline, (c) SwinIR~\cite{liang2021swinir}, (d) HAT~\cite{chen2023activating}, (e) SRFormer~\cite{zhou2023srformer}, (f) DAT~\cite{chen2023dual}, (g) CFAT~\cite{ray2024cfat}, (h) MambaIR~\cite{guo2024mambair}, (i) MambaIRv2~\cite{guo2025mambairv2}, (j) MaIR~\cite{li2025mair}, (k) ASSAN~\cite{li2025sparse}, and (l) the proposed ASBA. We further present the reconstructed B-scan visualizations under $\times 16$ sparsity on the MCD-AW and MCD-AN datasets in Figures~\ref{fig:ds16bscan-aw} and~\ref{fig:ds16bscan-an}. While reconstruction at $\times 8$ sparsity is more tractable than the more extreme $\times 16$ setting, it remains highly challenging, particularly for capturing slow blood flow in small vascular structures. Compared to other methods, which fail to recover many fine details under $\times 8$ and $\times 16$ sparsity, the proposed ASBA achieves successful reconstruction, further demonstrating its effectiveness in handling highly sparsely sampled raw ODT B-scans.

\end{document}